\documentclass[11pt]{article}

\usepackage{acl}

\usepackage{times}
\usepackage{latexsym}

\usepackage[T1]{fontenc}

\usepackage[utf8]{inputenc}

\usepackage{microtype}

\usepackage{inconsolata}

\usepackage{graphicx}

\usepackage{amsmath}
\usepackage{amssymb}

\usepackage{booktabs}

\title{AMU: Admission and Memory Update for Personalized Conversations — Structured Memory with SLM Guided Control}

\author{
\textbf{Tao Hwang}\textsuperscript{1,*},
\textbf{Yishi Diao}\textsuperscript{2}
\\[2pt]
\textsuperscript{1}Independent Researcher,
\textsuperscript{2}Nanchang University
\\[2pt]
\textbf{Correspondence:} \texttt{taohwang@ieee.org}
}

\begin{document}
\pagestyle{plain}
\maketitle
\thispagestyle{plain}

\begingroup
\renewcommand{\thefootnote}{*}
\footnotetext{Corresponding author.}
\endgroup

\begin{abstract}
Large language models (LLMs) have become the foundation of personalized assistants, but maintaining persistent user memory across long-term interactions remains challenging. Existing memory systems often focus on storage, retrieval, or consolidation, while memory writing remains less controlled: transient requests, duplicate statements, and outdated user states may enter memory and later be retrieved for personalization. In this paper, we present \textbf{AMU}: \textbf{A}dmission and \textbf{M}emory \textbf{U}pdate for Personalized Conversations, an SLM-guided (Small language model guided) structured framework for writing-time memory control. AMU uses structured memory filtering to decide what should enter memory and SLM-guided storage management to determine whether an admitted record should be stored separately, discarded as a duplicate, or fused as an update. We evaluate AMU in a controlled memory writing and retrieval setting. Experimental results show that AMU maintains cleaner and more retrievable personalized memories. Our code is available at \url{https://github.com/UnicusT11/AMU-memory}.
\end{abstract}

\section{Introduction}

Large language models (LLMs) are now widely used in conversational agents and personalized assistants. Recent models have shown strong abilities in instruction following, reasoning, tool use, multilingual interaction, and multimodal understanding \citep{NEURIPS2022_b1efde53, openai2024gpt4technicalreport, Wang_2024, geminiteam2025geminifamilyhighlycapable, grattafiori2024llama3herdmodels}. However, many LLM-based interactions are still organized around a single session. The model can respond to the current prompt, but it does not naturally maintain a stable and evolving understanding of the user across long-horizon interactions \citep{zhong2023memorybankenhancinglargelanguage, westhäußer2025enablingpersonalizedlongterminteractions}. For personalized assistants, this makes persistent memory necessary, especially for reusable user information such as preferences, background, plans, and changing personal states.

\begin{figure}[t]
  \centering
  \includegraphics[width=1.0\columnwidth]{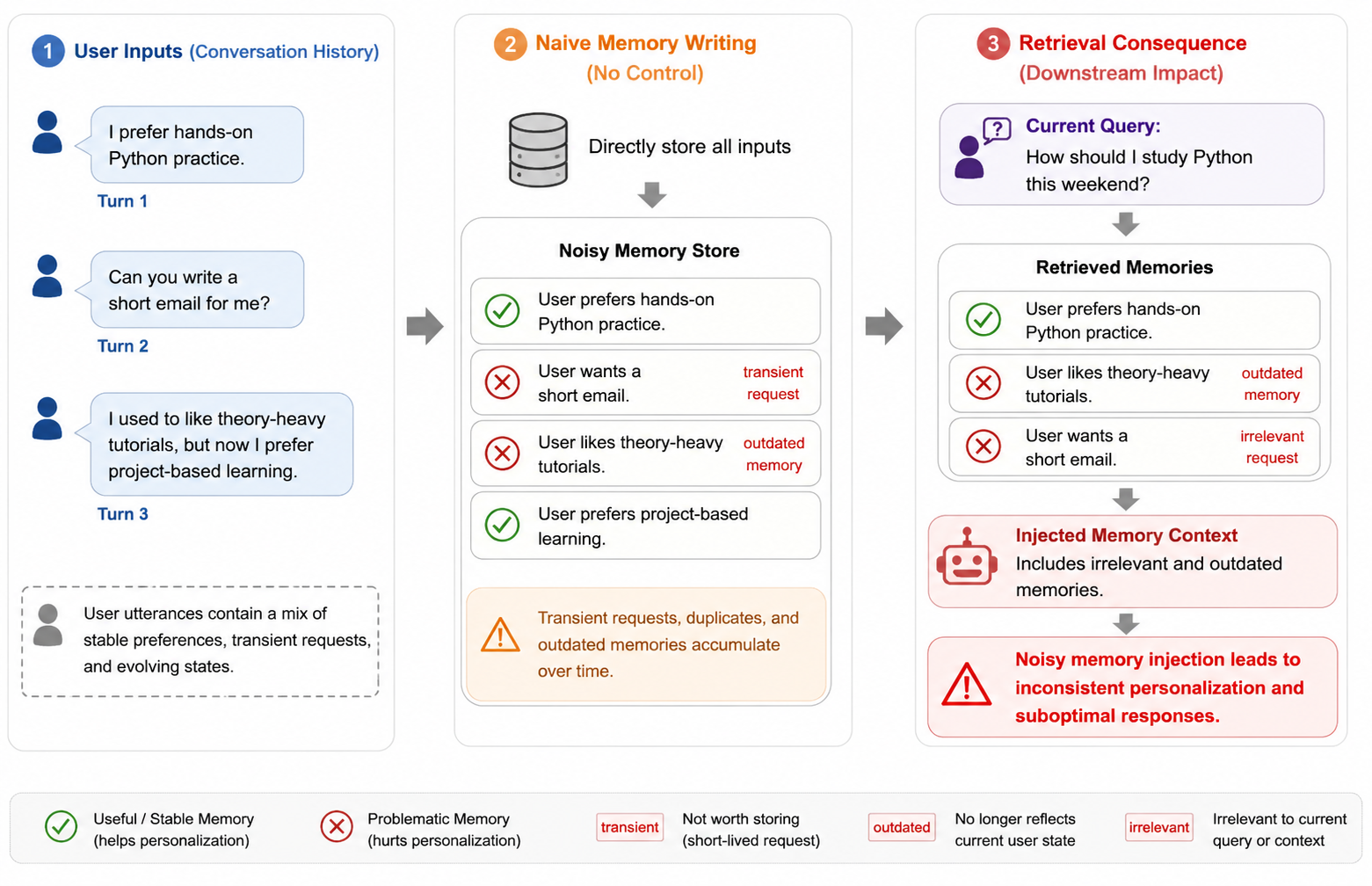}
  \caption{Motivation of writing-time memory control. Directly storing user utterances can preserve transient, redundant, or outdated memories, while AMU keeps stable and reusable memories for cleaner personalization.}
  \label{fig:intro}
\end{figure}

Existing work has supported such continuity through several directions. Long-context methods place more dialogue history into the prompt or train models to handle longer contexts \citep{bai-etal-2024-longbench, liu-etal-2024-lost, chen2024longlora}, but they can be costly and still do not ensure that the model uses the right evidence. Model-internal memory methods introduce memory tiers or read-write memory units inside the agent or model architecture \citep{packer2024memgptllmsoperatingsystems, modarressi2024retllmgeneralreadwritememory}, but these designs are often tied to particular implementations. External memory modules take a different route: they extract, store, retrieve, and inject user information outside the response model. This makes them practical for deployment, since the memory component can be updated without retraining and can be reused across different model backends \citep{zhou2026externalizationllmagentsunified, Rafique_2026}. Prior systems have explored a range of external-memory designs for storing, retrieving, and updating user information \citep{izacard-grave-2021-leveraging, 10.1145/3586183.3606763, li-etal-2025-hello, NEURIPS2025_19909c36, chhikara2025mem0buildingproductionreadyai}.

Despite these developments, fine-grained control at the memory writing stage remains underexplored. User conversations contain mixed signals. Some inputs describe stable preferences or long-term facts, while others are temporary requests, repeated statements, outdated states, emotional expressions, or details that are only weakly relevant to future personalization. If such information is admitted into memory without sufficient control, the memory store can gradually accumulate noisy, redundant, or outdated entries. Figure~\ref{fig:intro} illustrates this problem: transient requests and outdated preferences may be stored together with useful memories and later retrieved, which can lead to noisy memory injection and inconsistent personalization. A memory system therefore needs to decide not only which stored memories should be retrieved, but also which information should be written into memory in the first place and how it should interact with existing records.

Retrieval filtering or post-hoc consolidation only partly addresses this problem. Retrieval can select stored memories for a particular query, but it cannot remove the effect of low-value records that have already been written into the memory store. Consolidation can merge or revise stored memories, but when admission is weak, it may still operate over a noisy candidate pool. A robust personalized memory system therefore needs two coupled writing-time capabilities. The first is admission control, which filters low-value or non-personal inputs before storage. The second is relation-aware memory update, which determines whether an admitted record should be stored as a new memory, discarded as a duplicate, or fused into an updated user state. These capabilities are particularly important in long-term personalized settings, where preferences, plans, and personal states may change over repeated interactions.

In this paper, we propose \textbf{AMU}: \textbf{A}dmission and \textbf{M}emory \textbf{U}pdate for Personalized Conversations, a structured framework for writing-time memory control in personalized conversations. AMU focuses on the memory writing stage and targets two problems caused by passive memory accumulation: low-value or non-personal information may enter long-term memory, and repeated or evolving user states may create redundant or outdated records. In AMU, structured admission first determines which user inputs are eligible for memory storage. A small language model (SLM) is then used as a lightweight controller to decide how each admitted record should interact with existing memories. Specifically, the SLM judges whether a candidate memory should be stored separately, discarded as a duplicate, or fused as an update to the current reusable user state. The resulting memories are kept as structured records and serialized into plain-text memory context for downstream RAG-based response generation.

Our contributions are summarized as follows:
\begin{itemize}
    \item We introduce \textbf{AMU}, a structured framework for writing-time memory control in personalized conversations. AMU manages user memories before downstream retrieval by treating memory construction as a controlled writing process rather than passive accumulation.

    \item We propose structured admission for personalized memory writing. By parsing user inputs into typed memory records, AMU filters low-value or non-personal information before storage and uses structured fields to guide candidate selection and retrieval-time filtering.

    \item We propose SLM-guided memory update for maintaining reusable and evolving user memories. A lightweight SLM determines whether an admitted memory is duplicate, update, or separate from existing memories, and update-related records are fused into the current reusable user state.
\end{itemize}

\section{Related Work}
\label{app:related_work}

Large language models (LLMs) have enabled strong conversational and reasoning abilities, but persistent personalization remains difficult when user information must be maintained across long-term interactions. Existing personalized dialogue systems commonly condition responses on personas, user profiles, or dialogue histories to improve consistency and user adaptation \citep{ shuster2022blenderbot3deployedconversational}. Recent personalization benchmarks further evaluate whether LLMs can use user profiles, preferences, and historical interactions when generating responses \citep{salemi-etal-2024-lamp}. However, these studies mainly focus on how user information is used during generation, while the process by which user utterances are admitted, filtered, updated, or discarded before retrieval is less explicitly studied.

\paragraph{Personalized Memory.}
Early personalized dialogue work usually relies on static persona descriptions or profile sentences. Personal Chat \citep{zhang-etal-2018-personalizing} introduced persona-grounded conversations to evaluate whether models can produce responses consistent with assigned personas. \citet{dinan-etal-2020-queens} further studied controllable and persona-aware dialogue modeling in open-domain conversations. More recent work explores longer-term personalization and persistent memory management, where agents accumulate and maintain user-specific information from interaction histories rather than relying only on predefined persona statements. Mem0 maintains persistent memories through explicit memory operations, A-MAC focuses on adaptive memory admission and conflict-aware maintenance, and A-MEM organizes memories through structured linking and dynamic evolution \citep{chhikara2025mem0buildingproductionreadyai, zhang2026adaptivememoryadmissioncontrol, NEURIPS2025_19909c36}. These studies show that memory construction and maintenance are increasingly important for long-term personalized interaction. AMU focuses specifically on writing-time control for personalized conversational memory, coupling structured admission with relation-aware decisions over whether an admitted memory should be treated as duplicate, update, or separate.

\paragraph{Retrieval-Augmented Memory.}
Retrieval-augmented language models use external knowledge or non-parametric memory to improve factuality and task performance. REALM retrieves external knowledge during pre-training and downstream prediction \citep{pmlr-v119-guu20a}, and RETRO conditions generation on retrieved chunks from a large-scale retrieval database \citep{pmlr-v162-borgeaud22a}. Retrieval augmentation has also been applied to black-box and in-context language models, showing that retrieved evidence can improve model outputs without changing model parameters \citep{ram-etal-2023-context, shi2023replugretrievalaugmentedblackboxlanguage}. In dialogue and question answering, retrieval helps ground responses and reduce unsupported generation \citep{shuster-etal-2021-retrieval-augmentation, komeili-etal-2022-internet, nakano2022webgptbrowserassistedquestionansweringhuman}. More recent methods study adaptive or self-reflective retrieval, deciding when retrieval is needed and how evidence should be incorporated \citep{asai2024selfrag, jiang-etal-2023-active, mallen-etal-2023-trust}. However, these works mainly address how to retrieve and use stored information at inference time. They do not fully address how a personalized memory store should be kept clean before retrieval. AMU complements retrieval-augmented methods by applying structured filtering during memory writing and retrieval.

\paragraph{Memory Construction and Management.}
Agent-oriented methods have shown that intermediate reasoning traces, feedback, and memory-like states can improve long-horizon behavior. ReAct combines reasoning with external actions in an agent loop \citep{yao2023react}, Reflexion uses verbal feedback as an episodic memory signal for future decisions \citep{NEURIPS2023_1b44b878}, and Self-Refine iteratively improves model outputs through self-generated feedback \citep{NEURIPS2023_91edff07}. These methods demonstrate the usefulness of model-generated judgments and feedback. Nevertheless, personalized memory management requires a different form of decision-making: a system must decide whether a user utterance should be stored, discarded as redundant, or used to update an existing memory. AMU addresses this issue through structured memory filtering and SLM-guided storage management, where candidate memories are explicitly judged as duplicate, update, or separate before being written back to the memory store.

\section{Admission and Memory Update for Personalized Conversations}
\label{sec:amu}

This section describes AMU, an SLM-guided structured framework for personalized memory management. \S\ref{sec:amu_framework} presents the overall framework. \S\ref{sec:filtering} describes structured memory filtering, \S\ref{sec:storage_management} introduces SLM-guided storage management, and \S\ref{sec:storage_retrieval} describes structured storage and RAG-based retrieval.

\begin{figure*}[t]
  \centering
  \includegraphics[width=1.0\textwidth]{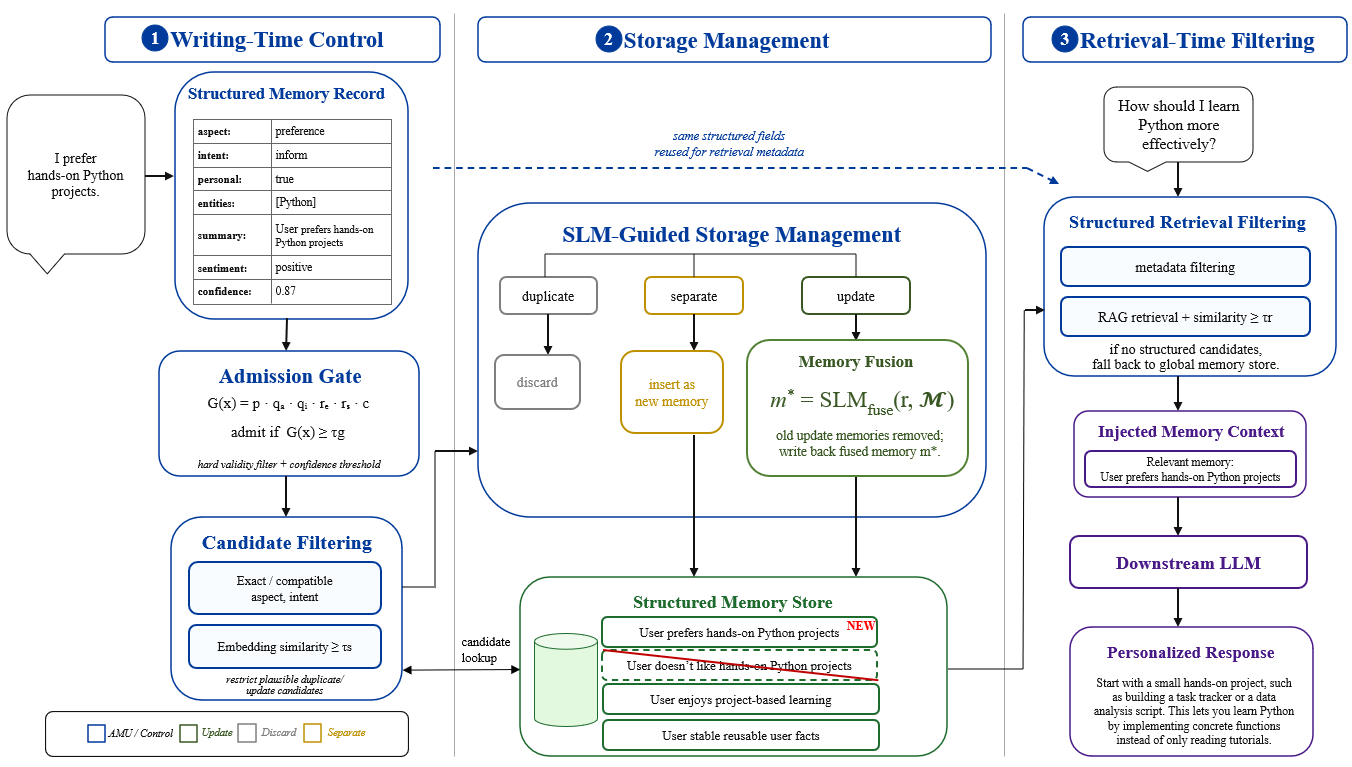}
  \caption{Overview of AMU. Structured fields guide memory admission and retrieval, while the SLM decides whether an admitted record is duplicate, update, or separate before storage.}
  \label{fig:framework}
\end{figure*}

\subsection{AMU Framework}

\label{sec:amu_framework}

AMU maintains personalized memories as structured and updateable records rather than an append-only log of user utterances. As shown in Figure~\ref{fig:framework}, AMU consists of two coupled mechanisms: structured memory filtering and SLM-guided storage management. Given a user input, AMU first converts it into a typed memory record. The structured fields are then used to determine whether the input should enter memory processing, which existing memories should be considered as candidates, and which memory subset should be exposed during retrieval. For admitted records, the SLM judges whether the new record is duplicate, update, or separate with respect to existing memories. Separate records are inserted, duplicate records are discarded, and update-related records are fused into the current reusable memory state. By decoupling memory management from the downstream response LLM, AMU supports early exits and enables the maintained memory store to be reused across different LLM backends.

\subsection{Structured Memory Filtering}

\label{sec:filtering}

\paragraph{Structured memory record.}

Structured memory filtering begins with a typed memory record. Given a user input, we use an SLM to produce

\begin{equation}
m = \{a, i, p, e, s, v, c\}
\end{equation}

where \(a\) denotes the memory aspect, \(i\) denotes the communicative intent, \(p\) indicates reusable personal information, \(e\) contains salient entities, \(s\) is an affect-neutral memory summary, \(v\) is sentiment, and \(c\) is confidence. The parsing process first extracts structural fields, including aspect, intent, reusable-personal flag, and entities, and then formulates the memory summary, sentiment, and confidence score. This step constructs the representation used by subsequent filtering and storage management.

Each user input is structured with two complementary labels: aspect for the type of user information and intent for the communicative function of the utterance. Aspect labels correspond to common facets in user modeling and personal information understanding, such as profile, relationship, activity, health, and preference \citep{purificato2024usermodelinguserprofiling, tan-etal-2025-personabench}, while intent labels correspond to dialogue-act-style abstraction of utterance function \citep{vasselli-etal-2025-multilingual}. Accordingly, we define the aspect set as \{\texttt{personal\_profile}, \texttt{social\_relationship}, \texttt{activity}, \texttt{health}, \texttt{preference}, \texttt{other}\}, and the intent set as \{\texttt{inform}, \texttt{request}, \texttt{express}, \texttt{plan}, \texttt{reflect}, \texttt{other}\}. These labels provide complementary structural views: aspect captures what the memory is about, while intent captures how the user presents it. The entity field anchors memory formulation but is not used as a hard matching key.

The memory summary \(s\) is stored as an affect-neutral statement that preserves one explicit, stable, and reusable user fact, while sentiment \(v\) is stored separately as \{\texttt{positive}, \texttt{neutral}, \texttt{negative}\}. This separation keeps factual memory content distinct from affective tone. We use the confidence score \(c\) as a lightweight reliability signal for the generated memory record, corresponding to calibration and self-evaluation in long-form generation \citep{huang-etal-2024-calibrating}. Since model-generated confidence is not necessarily calibrated, \(c\) is elicited with a rubric over explicitness, stability, and ambiguity, rather than treated as an unconstrained probability estimate.

\paragraph{Admission gate.}

Based on the structured record, an admission gate determines whether the record enters the memory writing pipeline. We formulate memory admission as a writing-time control decision based on content type and persistence signals in memory admission control \citep{zhang2026adaptivememoryadmissioncontrol}. Given \(m=\{a,i,p,e,s,v,c\}\), the gate computes

\begin{equation}
G(x)=p\cdot q_a\cdot q_i\cdot r_e\cdot r_s\cdot c
\end{equation}

Here, \(p\) indicates reusable personal information, \(q_a\) filters fallback aspects, \(q_i\) filters fallback intents and purely task-oriented requests without reusable personal content, and \(r_e\) and \(r_s\) indicate whether entities and memory summary are non-empty. Since all terms except \(c\) are binary, the gate implements a hard validity filter combined with a confidence threshold. A record is admitted only when \(G(x)\geq\tau_g\).

\paragraph{Candidate filtering.}

For admitted records, structured labels restrict the candidate memory pool. Existing memories with the same aspect-intent pair are first retrieved. Because these labels are predicted fields, the lookup expands to compatible intents under the same aspect when no exact-pair candidates are found. The resulting candidate set is further filtered by embedding similarity:

\begin{equation}
\mathrm{sim}_j=\mathrm{sim}(u,u_j), \quad \mathrm{sim}_j \geq \tau_s
\end{equation}

where \(u\) and \(u_j\) are embeddings of the current record and candidate memory \(m_j\). The threshold \(\tau_s\) is used only during writing for duplicate/update candidate filtering. This step limits later SLM judgment to plausible duplicate or update candidates.

\subsection{SLM-Guided Storage Management}

\label{sec:storage_management}

After structured filtering, we perform SLM-guided storage management over the retained candidates. For each candidate, a compact comparison view is constructed using entities, memory summary, and sentiment. The SLM predicts the storage relation between the current record and each candidate as one of three labels: \texttt{duplicate}, \texttt{update}, or \texttt{separate}.

A \texttt{duplicate} record adds no reusable information beyond an existing memory and is discarded. An \texttt{update} record revises, narrows, contradicts, or supersedes an existing memory and should replace the old state. A \texttt{separate} record describes another stable user fact and can coexist with existing memories. This relational judgment turns memory writing from a binary store-or-drop decision into a storage management process that supports insertion, deletion, and update.

Deterministic actions are applied according to the predicted relations. If all candidates are separate, the current record is inserted as a new memory. If no update is found but at least one duplicate exists, the current record is discarded as redundant. If one or more update candidates are found, memory fusion is performed. We treat update as a storage-management relation for evolving user states, where outdated preferences or states should be revised rather than kept active \citep{sun2025preferenceawarememoryupdatelongterm}. Let \(d\) denote the current structured record and \(\mathcal{M}\) denote the list of candidate memories predicted as \texttt{update}. The SLM fuses \(d\) and \(\mathcal{M}\) into a new memory record:
\[
m^{*}=\mathrm{SLM}_{\mathrm{fuse}}(d,\mathcal{M}).
\]
Only update candidates are used for fusion; separate candidates remain unchanged, and duplicates are excluded. The fused memory \(m^*\) represents the current reusable state. Old update memories are removed, and only \(m^*\) is written back to the memory store.

\subsection{Structured Storage and RAG Retrieval}

\label{sec:storage_retrieval}

Accepted memories are stored as document-style structured records, with \texttt{memory\_summary} used as the main retrieval text for vector indexing. Structured fields are also used as retrieval metadata to filter candidate memories before vector retrieval, so retrieval is performed over a more relevant subset of the memory store. If no structured candidates are available, retrieval falls back to the global memory store.

During retrieval, the RAG backend uses the current user query to retrieve relevant memories and injects them as plain-text memory context. We further filter the retriever's top-\(k\) results with a retrieval-time threshold \(\tau_r\), which suppresses weakly related memories during inference. This threshold differs from \(\tau_s\), which is used during writing for duplicate/update detection. Thus, structured filtering is used at both writing time and retrieval time: writing-time filtering controls which records enter and update the memory store, while retrieval-time filtering controls which maintained memories are exposed for downstream use.

\section{Experiment}

\subsection{Dataset and Experiment Setup}
\label{sec:dataset_setup}

\paragraph{Dataset.}
We use GPT-4o to construct a controlled benchmark for memory writing and retrieval by adapting LoCoMo-style long-term conversational contexts into single-topic sessions with explicit action labels and retrieval targets \citep{maharana-etal-2024-evaluating, kim-etal-2025-evaluating}. This setup is motivated by the privacy sensitivity and annotation difficulty of real long-term personal memory logs. The benchmark simulates one to two months of theme-based interactions, including stable preferences, evolving attitudes, repeated information, and preference or plan updates. Each session is generated under a specific theme with a guideline-constrained prompt and a fixed JSON schema. We generate 20 sessions with different themes, and each session contains 50 memory-writing turns followed by 10 evaluation queries. The benchmark contains 1,000 memory-writing turns and 200 evaluation queries in total. Each memory-writing turn is annotated with a construction-time action label, including \texttt{store}, \texttt{discard}, \texttt{update}, or \texttt{filter}, and each memory-bearing turn is assigned a slot-level \texttt{memory\_id}. Each evaluation query is associated with expected memory ids for retrieval evaluation.

\paragraph{Experimental setting.}
Experiments are conducted on the 20 sessions. For each session, the first 50 turns are processed sequentially as memory-writing inputs, and the last 10 queries evaluate retrieval against the expected slot-level memory ids. All memory-based systems use the same Haystack-based RAG backend and Qwen3-Embedding-0.6B embedding model. The compared systems differ in how memories are constructed, admitted, updated, and maintained, while the downstream retrieval protocol is kept fixed for controlled evaluation. Qwen3.5-2B is used as the default SLM for memory management unless otherwise specified. All experiments were conducted on a single NVIDIA RTX 4070 GPU.

\subsection{Manual Validation of Silver Labels}
\label{sec:Manual Validation of Silver Labels}

To assess the reliability of GPT-4o-generated silver annotations, the same subset of writing turns and evaluation queries was independently reviewed by multiple annotators. The subset was selected through programmatic random sampling of item IDs and contained 30 writing turns and 5 queries from each session, corresponding to 600 of 1,000 writing turns (60\%) and 100 of 200 evaluation queries (50\%). For writing turns, the annotators judged whether the assigned action matched the definitions of \texttt{store}, \texttt{discard}, \texttt{update}, and \texttt{filter}. For evaluation queries, they verified whether the expected \texttt{memory\_ids} were traceable to the corresponding writing turns and appropriate for the query.

Table~\ref{tab:annotation_validity} reports the mean validity rates across the independent annotations: 89.3\% for action-label validity and 94.1\% for query-memory alignment. These values measure manual annotation validity rather than inter-annotator agreement.

\begin{table}[htbp]
  \centering
  \small
  \begin{tabular}{lc}
    \toprule
    \textbf{Metric} & \textbf{Validity (\%)} \\
    \midrule
    Action label validity & 89.3 \\
    Query-memory alignment & 94.1 \\
    \bottomrule
  \end{tabular}
  \caption{Manual validation of silver annotations. Reported values are mean validity rates across independent annotations.}
  \label{tab:annotation_validity}
\end{table}

\begin{table*}[t]
\centering
\normalsize
\setlength{\tabcolsep}{8pt}
\begin{tabular}{lccccc}
\toprule
\textbf{Method} & \textbf{P@3} & \textbf{R@3} & \textbf{F1@3} & \textbf{HIT@3} & \textbf{Red.@3 $\downarrow$} \\
\midrule
Full Comp. & 35.42 & 55.50 & 41.30 & 57.00 & 12.08 \\
Sliding Window & 24.42 & 29.50 & 25.87 & 29.50 & 11.75 \\
Mem0 & 34.08 & 57.50 & 40.88 & 56.00 & 3.63 \\
A-MAC & 30.36 & 58.76 & 37.96 & 54.50 & 3.18 \\
A-MEM & 35.13 & 56.90 & 40.36 & 56.50 & 2.88 \\
\textbf{AMU(Ours)} & \textbf{37.33} & \textbf{63.50} & \textbf{44.47} & \textbf{57.50} & \textbf{2.83} \\
\bottomrule
\end{tabular}
\caption{Main retrieval results. Red.@3 denotes Redundancy@3, where lower values are better.}
\label{tab:main_results}
\end{table*}

\subsection{Experimental Results}
\label{sec:main_results}

\paragraph{Compared baselines.}
We compare AMU with five baselines covering both simple memory-retention strategies and existing long-term memory-management methods. Full Compressed Memory compresses and stores all memory-writing inputs, representing summary- or compression-based memory construction in long-term memory systems \citep{chen-etal-2025-compress, zhang2026lightweightllmagentmemory}. Sliding Window retains only recent memories and discards older ones, providing a recency-based reference under limited memory capacity. Mem0 \citep{chhikara2025mem0buildingproductionreadyai} extracts salient information from conversational history and maintains persistent memories through explicit memory operations, including addition, update, deletion, and no-op. A-MAC \citep{zhang2026adaptivememoryadmissioncontrol} performs adaptive memory admission using signals such as utility, confidence, novelty, recency, and content type, together with conflict-aware memory maintenance, providing a direct comparison for selective memory writing and maintenance. A-MEM \citep{NEURIPS2025_19909c36} constructs structured memory notes and dynamically links and evolves stored memories as new information is incorporated. All methods are evaluated under the same slot-level retrieval protocol to compare the quality of the maintained memory stores.

\paragraph{Evaluation metrics.}
We evaluate top-3 retrieval on memory-required queries. Retrieved memories are mapped to slot-level \texttt{memory\_id}s and compared with expected memory ids. We report P@3, R@3, F1@3, and HIT@3, measuring precision, recall, precision-recall balance, and whether at least one expected memory slot is retrieved. Since AMU targets cleaner memory maintenance rather than retrieval coverage alone, we also report Redundancy@3, defined as the percentage of top-3 retrieved memories that repeat an already retrieved slot-level \texttt{memory\_id}. Lower values indicate cleaner retrieval. No-memory queries are used only as a sanity check.

\paragraph{Results.}
As shown in Table~\ref{tab:main_results}, AMU achieves the best overall retrieval performance while maintaining the lowest redundancy. It obtains 37.33 P@3, 63.50 R@3, 44.47 F1@3, and 57.50 HIT@3, with a Red.@3 of 2.83. Compared with Full Compressed Memory, AMU improves both precision and recall while reducing Red.@3 from 12.08 to 2.83. This indicates that retaining all compressed memories preserves useful information but also introduces substantial redundant memory traces. Sliding Window performs substantially worse across retrieval metrics, showing that recency-only retention can discard information required for long-term personalization.
Among the existing memory-management methods, different approaches exhibit different retrieval--redundancy trade-offs. Mem0 achieves 57.50 R@3 and 40.88 F1@3 while substantially reducing redundancy relative to Full Compressed Memory. AMU further improves its P@3, R@3, and F1@3, while reducing Red.@3 from 3.63 to 2.83. A-MAC obtains the highest recall among the compared existing methods at 58.76, but its lower precision of 30.36 results in an F1@3 of 37.96. AMU improves both precision and recall over A-MAC, indicating a stronger precision–recall balance under our controlled memory-writing setting. A-MEM achieves a Red.@3 of 2.88, which is already close to AMU, but AMU increases R@3 from 56.90 to 63.50 and F1@3 from 40.36 to 44.47. This result is particularly important because AMU's improvement cannot be explained simply by producing a less redundant memory store.
Overall, the results show that AMU provides a stronger balance between retrieval effectiveness and memory cleanliness. Existing methods respectively provide explicit memory operations, selective admission, or structured memory evolution, yet AMU achieves the highest P@3, R@3, F1@3, and HIT@3 together with the lowest Red.@3 in our controlled setting. These results support combining structured admission with relation-aware duplicate, update, and separate decisions for maintaining reusable personalized memory before downstream retrieval.

\subsection{Component Verification}
\label{sec:component_verification}

To verify AMU's two main mechanisms, we conduct ablations with two variants. \textbf{AMU w/o Structured Filtering} removes structured filtering from memory writing and retrieval. \textbf{AMU w/o SLM Storage Management} keeps structured filtering but removes SLM-guided relation judgment, so \texttt{duplicate}, \texttt{update}, and \texttt{separate} relations are no longer explicitly handled. We report F1@3, Redundancy@3, and \textbf{Action Acc.}, defined as the percentage of memory-writing turns whose predicted action matches the gold action label.

As shown in Table~\ref{tab:ablation}, removing either mechanism weakens AMU. Without structured filtering, both F1@3 and Action Acc. decrease, indicating weaker noise filtering and retrieval alignment. Without SLM storage management, F1@3 and Action Acc. drop more substantially, showing that explicit duplicate/update/separate judgment is important for memory maintenance. Although this variant has lower Redundancy@3, the low redundancy is accompanied by poor retrieval coverage and unreliable writing decisions. These results show that structured memory filtering and SLM-guided storage management are complementary: the former selects records and candidates, while the latter decides how admitted memories are stored, discarded, or updated.

\begin{table}[t]
\centering
\small
\setlength{\tabcolsep}{3.5pt}
\begin{tabular}{lccc}
\hline
\textbf{Variant} & \textbf{F1@3} & \textbf{Red.@3 $\downarrow$} & \textbf{Act. Acc.} \\
\hline
w/o Struct. Filtering & 34.30 & 2.25 & 65.40 \\
w/o SLM Storage Mgmt. & 25.62 & 0.25 & 34.50 \\
AMU(Ours) & 44.47 & 2.83 & 67.70 \\
\hline
\end{tabular}
\caption{Component verification of AMU. Red.@3 denotes Redundancy@3, where lower values are better.}
\label{tab:ablation}
\end{table}

\section{Analysis}
\label{sec:analysis}

We analyze how the size of the SLM controller affects memory maintenance, focusing on the relation between writing-action accuracy and retrieval-side behavior. As shown in Table~\ref{tab:controller_size}, larger controllers improve Action Acc., especially from 0.8B to larger models, indicating that overly small controllers are not stable enough for structured parsing and storage decisions. However, after 4B, the accuracy gain becomes limited, while F1@3 and Red.@3 do not improve proportionally. This suggests that stronger writing-action prediction does not necessarily lead to better retrieval-side performance, since retrieval is also affected by embedding similarity, thresholds, and top-\(k\) ranking. The results therefore indicate limited marginal gains from further scaling in this setting, while showing that Qwen3.5-2B is sufficient for AMU without relying on a large response LLM for memory control.

\begin{table}[t]
\centering
\small
\begin{tabular}{lccc}
\hline
\textbf{Size} & \textbf{F1@3} & \textbf{Red.@3 $\downarrow$} & \textbf{Act. Acc.} \\
\hline
0.8B & 13.90 & 0.31 & 26.40 \\
2B   & 44.47 & 2.83 & 67.70 \\
4B   & 41.78 & 3.08 & 81.00 \\
9B   & 40.00 & 3.33 & 84.20 \\
\hline
\end{tabular}
\caption{Effect of SLM controller size.}
\label{tab:controller_size}
\end{table}

\section{Conclusion}

In this paper, we presented AMU: Admission and Memory Update for Personalized Conversations. AMU combines structured memory filtering with SLM-guided storage management. To conduct the evaluation, we construct a controlled memory writing and retrieval benchmark to measure whether systems can retrieve expected memory slots while avoiding redundant memories. In addition, we verify the potential of these components for improving memory maintenance. The results demonstrate the value of combining structured memory filtering with SLM-guided storage management for maintaining cleaner personalized memories before downstream retrieval and response generation.

\section*{Limitations of the Work}

\paragraph{Hardware-Dependent Running Time.}
All experiments are conducted on a single NVIDIA RTX 4070 GPU, representing an accessible local evaluation setting rather than a large-scale serving environment. This hardware setting provides a practical reference for evaluating AMU under local deployment constraints. However, the running time reported in our experiments should be interpreted together with this setup, since larger SLM controllers may run faster on stronger accelerators or optimized inference infrastructure.

\paragraph{SLM Guidance Dependence.}
AMU relies on an SLM to extract structured memory records and perform storage-management judgments. Its effectiveness therefore depends on whether the chosen SLM can follow structured prompts, produce stable fields, and distinguish \texttt{duplicate}, \texttt{update}, and \texttt{separate} relations. Very small or poorly aligned SLMs may generate unstable structures or unreliable relation judgments, which can affect both memory writing and downstream retrieval.

\section*{Declaration on Generative AI}
Generative AI tools were used solely for language polishing during manuscript preparation, including improvements to grammar, and clarity. 

\section*{Ethics Statement}
This work uses a controlled benchmark constructed for memory writing and retrieval. The benchmark is generated and transformed with GPT-4o by adapting textual contexts from LoCoMo-style long-term conversations into AMU's memory-writing setting. It does not rely on real user conversations or personally identifiable private data. The AMU-specific annotations, including memory-writing action labels, memory identifiers, update links, and retrieval targets, are created for controlled evaluation.

AMU is designed for personalized memory management, where stored memories may reflect user preferences, plans, or states. In practical deployment, users should be informed when memory is enabled and should be able to inspect, edit, or delete stored memories. Since AMU stores memories as structured records, selected memory entries can be removed or updated when users request deletion or correction.

\bibliography{amu_references}

\clearpage
\appendix
\raggedbottom
\makeatletter
\setlength{\@dblfptop}{0pt}
\setlength{\@dblfpsep}{8pt plus 1pt minus 1pt}
\setlength{\@dblfpbot}{0pt plus 1fil}
\makeatother

\setlength{\dbltextfloatsep}{8pt plus 1pt minus 2pt}
\setlength{\textfloatsep}{8pt plus 1pt minus 2pt}
\section*{Appendix}

\section{Additional Analysis}
\label{app:analysis}

We provide more details on the behavior of AMU under different settings. The additional analyses include downstream LLM use with different memory settings, sensitivity to writing-time and retrieval-time similarity thresholds, comparisons of embedding models, comparisons of different SLM backbones for memory management, and session mixing analysis with memories from multiple sessions.

\subsection{Downstream LLM Judge Evaluation}
\label{app:llm_judge}

To examine how the maintained memory affects downstream personalized responses, we conduct a response-level evaluation on sampled cases from each session. For each session, we construct the memory store from the memory-writing turns and randomly sample 5 evaluation queries for response generation. Responses are generated under three settings: without memory, with full compressed memory, and with AMU-maintained memory. The response model is Gemma3-12B with quantized inference, and AMU uses Qwen3.5-2B as the default SLM for memory management. GPT-4o is used as the judge. For each case, the judge is given the user query, the reference memory, the expected behavior, and the assistant response, and assigns an overall score from 1 to 5 based on memory use, avoidance of irrelevant or outdated personal information, and helpfulness.

As shown in Table~\ref{tab:llm_judge}, AMU receives the highest overall score. No Memory lacks access to relevant user information while Full Compressed Memory provides only a modest improvement, suggesting that unfiltered compressed memories may fail to provide useful personalization when irrelevant or redundant information is included. AMU provides a cleaner maintained memory state, allowing the downstream LLM to use relevant personal information more effectively. This analysis is intended as a small-scale response-level supplement rather than a full end-to-end personalization benchmark.

\begin{table}[htbp]
\centering
\small
\begin{tabular}{lc}
\hline
\textbf{Method} & \textbf{Overall Score} \\
\hline
No Memory & 2.92 \\
Full Compressed Memory & 3.21 \\
AMU & \textbf{4.37} \\
\hline
\end{tabular}
\caption{Small-scale downstream response evaluation with an LLM judge.}
\label{tab:llm_judge}
\end{table}

\subsection{Threshold Sensitivity}
\label{app:threshold}

We analyze the sensitivity of AMU to the writing-time similarity threshold \(\tau_s\) and the retrieval-time similarity threshold \(\tau_r\). We set \(\tau_s=0.60\) and \(\tau_r=0.55\) as the default thresholds, and vary one threshold at a time to examine its effect on memory writing, retrieval quality, redundancy, and efficiency.

\begin{table}[htbp]
\centering
\footnotesize
\setlength{\tabcolsep}{3.2pt}
\begin{tabular}{ccccc}
\hline
\(\boldsymbol{\tau_s}\) & \textbf{F1@3} & \textbf{Red.@3 \(\downarrow\)} & \textbf{Acc.} & \textbf{Time/Turn \(\downarrow\)} \\
\hline
0.40 & 43.68 & 2.75 & 66.80 & 327.17 \\
0.60 & 44.47 & 2.83 & 67.70 & 231.34 \\
0.80 & 45.82 & 4.50 & 65.80 & 176.74 \\
\hline
\end{tabular}
\caption{Sensitivity to the writing-time threshold \(\tau_s\) with \(\tau_r=0.55\). Red.@3 denotes Redundancy@3, where lower values are better.}
\label{tab:threshold_writing}
\end{table}

\begin{table}[htbp]
\centering
\footnotesize
\setlength{\tabcolsep}{3.0pt}
\begin{tabular}{cccccc}
\hline
\(\boldsymbol{\tau_r}\) & \textbf{P@3} & \textbf{R@3} & \textbf{F1@3} & \textbf{Red.@3 \(\downarrow\)} & \textbf{HIT@3} \\
\hline
0.40 & 36.00 & 64.25 & 43.75 & 2.83 & 58.00 \\
0.55 & 37.33 & 63.50 & 44.47 & 2.83 & 57.50 \\
0.70 & 45.33 & 51.25 & 46.33 & 2.00 & 44.50 \\
\hline
\end{tabular}
\caption{Sensitivity to the retrieval-time threshold \(\tau_r\) with \(\tau_s=0.60\). Red.@3 denotes Redundancy@3, where lower values are better.}
\label{tab:threshold_retrieval}
\end{table}

Table~\ref{tab:threshold_writing} shows the effect of varying \(\tau_s\) while fixing \(\tau_r=0.55\). A smaller \(\tau_s\) retains more candidate memories for SLM-guided storage management, which increases the average processing time per turn. As \(\tau_s\) increases, fewer candidates enter relation judgment, reducing Time/Turn. However, a larger \(\tau_s\) also leaves fewer opportunities for duplicate or update management, which can increase redundancy. In this setting, \(\tau_s=0.60\) provides the highest Action Acc. and a balanced trade-off between writing reliability, retrieval performance, redundancy, and efficiency.

Table~\ref{tab:threshold_retrieval} shows the effect of varying \(\tau_r\) while fixing \(\tau_s=0.60\). Increasing \(\tau_r\) makes retrieval-time filtering stricter, which improves P@3 and Redundancy@3 by suppressing weakly related memories. However, the stricter threshold also reduces R@3 and HIT@3, indicating that some useful memories may be filtered out. Thus, \(\tau_r\) mainly controls the precision-recall trade-off during retrieval. We use \(\tau_r=0.55\) as the default retrieval-time threshold because it preserves higher recall and HIT@3 while maintaining low redundancy.

\subsection{Embedding Model Comparison}
\label{app:embedding}

We compare three embedding models used in both memory writing and retrieval: Qwen3-Embedding-0.6B, BGE-M3, and Multilingual-E5-Large. For each setting, the same embedding model is used for writing-time candidate filtering and retrieval-time memory retrieval.

As shown in Table~\ref{tab:additional_analysis}(a), changing the embedding model leads to only small differences across retrieval and writing-related metrics. The three embedding models produce close P@3, R@3, F1@3, HIT@3, Redundancy@3, and Action Acc. scores, indicating that replacing the embedding model in both writing-time candidate filtering and retrieval-time memory retrieval does not substantially alter AMU's behavior. This suggests that AMU is not tightly coupled to a specific retriever encoder, and the structured memory filtering and SLM-guided storage management pipeline remains stable across the tested embedding choices.

\subsection{SLM Choice for Storage Management}
\label{app:slm_choice}

We compare different SLMs used for memory management to examine whether AMU is sensitive to the choice of the SLM backbone. The compared models include Qwen3.5-2B, Qwen3-4B, Gemma-2B-IT, and Gemma4-E2B-IT. All other components, including the embedding model, thresholds, and retrieval pipeline, are kept unchanged.

As shown in Table~\ref{tab:additional_analysis}(b), changing the SLM backbone leads to moderate differences but does not substantially change the behavior of AMU. Models with larger parameter sizes generally provide stronger results, especially in writing-action accuracy and retrieval metrics, while smaller models show slightly weaker performance. However, the differences across most settings remain within a limited range, suggesting that AMU is not tightly coupled to a specific SLM backbone. The structured memory filtering and SLM-guided storage management pipeline remains effective across the tested SLM choices.

\subsection{Session Mixing Analysis}
\label{app:session_mixing}

We further analyze whether AMU is affected when memories from different sessions are mixed in the same memory store. Sessions are randomly sampled to construct single-session, two-session, and three-session memory stores. Each setting is repeated over five random draws, and the reported scores are averaged across runs.

As shown in Table~\ref{tab:additional_analysis}(c), AMU remains stable when memories from multiple sessions are mixed in the same memory store. The retrieval metrics stay close across the single-session and mixed-session settings, and no consistent degradation is observed as the number of mixed sessions increases. This suggests that AMU can still select relevant memories from a moderately mixed memory store, which is consistent with the role of structured filtering and storage management.

\begin{table*}[!t]
\centering
\small
\caption{Additional analysis of embedding models, SLM backbones, and session mixing. Red.@3 and Red. denote Redundancy@3, where lower values are better.}
\label{tab:additional_analysis}
\vspace{-0.4em}

\begin{minipage}{0.72\textwidth}
\centering
\textbf{(a) Effect of different embedding models.}

\vspace{2pt}
\begin{tabular*}{\linewidth}{@{\extracolsep{\fill}}lcccccc@{}}
\hline
\textbf{Embedding Model} & \textbf{P@3} & \textbf{R@3} & \textbf{F1@3} & \textbf{Red.@3 $\downarrow$} & \textbf{HIT@3} & \textbf{Action Acc.} \\
\hline
Qwen3-Embedding-0.6B & 37.33 & 63.50 & 44.47 & 2.83 & 57.50 & 67.70 \\
BGE-M3 & 36.83 & 63.00 & 43.97 & 2.92 & 57.00 & 67.10 \\
Multilingual-E5-Large & 36.58 & 62.75 & 43.68 & 2.75 & 56.50 & 66.90 \\
\hline
\end{tabular*}
\end{minipage}

\vspace{0.7em}

\begin{minipage}{0.72\textwidth}
\centering
\textbf{(b) Effect of different SLM backbones used for memory management.}

\vspace{2pt}
\begin{tabular*}{\linewidth}{@{\extracolsep{\fill}}lcccccc@{}}
\hline
\textbf{SLM Backbone} & \textbf{P@3} & \textbf{R@3} & \textbf{F1@3} & \textbf{Red.@3 $\downarrow$} & \textbf{HIT@3} & \textbf{Action Acc.} \\
\hline
Qwen3.5-2B & 37.33 & 63.50 & 44.47 & 2.83 & 57.50 & 67.70 \\
Qwen3-4B & 37.33 & 66.00 & 45.27 & 2.83 & 59.00 & 78.70 \\
Gemma-2B-IT & 30.50 & 55.00 & 37.22 & 8.25 & 55.00 & 65.70 \\
Gemma4-E2B-IT & 37.67 & 66.75 & 45.73 & 1.67 & 61.50 & 64.10 \\
\hline
\end{tabular*}
\end{minipage}

\vspace{0.7em}

\begin{minipage}{0.72\textwidth}
\centering
\textbf{(c) Effect of mixing memories from different sessions.}

\vspace{2pt}
\begin{tabular*}{\linewidth}{@{\extracolsep{\fill}}lccccc@{}}
\hline
\textbf{Setting} & \textbf{P@3} & \textbf{R@3} & \textbf{F1@3} & \textbf{Red. $\downarrow$} & \textbf{HIT@3} \\
\hline
Single & 38.68 & 69.00 & 47.12 & 4.34 & 64.00 \\
2-mix & 38.00 & 66.00 & 45.98 & 2.66 & 60.00 \\
3-mix & 42.52 & 69.00 & 47.14 & 4.00 & 64.00 \\
\hline
\end{tabular*}
\end{minipage}

\end{table*}

\end{document}